\documentclass[conference]{IEEEtran}
\IEEEoverridecommandlockouts
\usepackage{cite}
\usepackage{amsmath,amssymb,amsfonts}
\usepackage{algorithmic}
\usepackage{graphicx}
\usepackage{textcomp}
\usepackage{xcolor}
\def\BibTeX{{\rm B\kern-.05em{\sc i\kern-.025em b}\kern-.08em
    T\kern-.1667em\lower.7ex\hbox{E}\kern-.125emX}}

\usepackage{hyperref}

\usepackage{tikz}
\usetikzlibrary{arrows,arrows.meta,shapes,positioning,shadows,trees,calc,decorations.pathmorphing}
\tikzset{
	basic/.style  = {draw, text width=2cm, drop shadow, font=\sffamily, rectangle},
	root/.style   = {basic, rounded corners=2pt, thin, align=center,
		fill=green!30},
	level 2/.style = {basic, rounded corners=6pt, thin,align=center, fill=green!60,
		text width=8em},
	level 3/.style = {basic, thin, align=center, fill=pink!60, text width=9.5em}
}

\begin{document}

\title{WRDScore: New Metric for Evaluation of Natural Language Generation Models\\
\thanks{This research was funded by the Science Committee of the Ministry of Science and Higher Education of the Republic of Kazakhstan (grant no. BR21882268).}
}

\author{\IEEEauthorblockN{Ravil Mussabayev}
\IEEEauthorblockA{\textit{AI Research Laboratory} \\
\textit{Satbayev University}\\
Almaty, Kazakhstan \\
r.mussabayev@satbayev.university}
}

\maketitle

\begin{abstract}
Evaluating natural language generation models, particularly for method name prediction, poses significant challenges. A robust metric must account for the versatility of method naming, considering both semantic and syntactic variations. Traditional overlap-based metrics, such as ROUGE, fail to capture these nuances. Existing embedding-based metrics often suffer from imbalanced precision and recall, lack normalized scores, or make unrealistic assumptions about sequences. To address these limitations, we leverage the theory of optimal transport and construct WRDScore, a novel metric that strikes a balance between simplicity and effectiveness. In the WRDScore framework, we define precision as the maximum degree to which the predicted sequence's tokens are included in the reference sequence, token by token. Recall is calculated as the total cost of the optimal transport plan that maps the reference sequence to the predicted one. Finally, WRDScore is computed as the harmonic mean of precision and recall, balancing these two complementary metrics. Our metric is lightweight, normalized, and precision-recall-oriented, avoiding unrealistic assumptions while aligning well with human judgments. Experiments on a human-curated dataset confirm the superiority of WRDScore over other available text metrics.
\end{abstract}

\begin{IEEEkeywords}
Natural language generation, Method name prediction, AI for code, Word Mover's Distance
\end{IEEEkeywords}

\section{Introduction}

The growing adoption of deep-learning-based natural language generation models has necessitated the development of advanced evaluation metrics. Existing metrics fall short in capturing subtle semantic and syntactic differences between generated and reference texts. Traditional metrics like ROUGE \cite{Lin2004} merely measure token overlap, neglecting essential aspects such as paraphrasing, synonym replacement, abbreviations, and token order. For example, ROUGE-1 fails to recognize the clear similarity between words like ``copy'' and ``clone'', or ``size'' and ``get count'', highlighting the need for more sophisticated evaluation methods.

In this paper, we propose a novel metric for evaluating natural language generation (NLG) model outputs, specifically designed for Java method name prediction. Existing embedding-based NLG metrics, such as BERTScore \cite{Zhang2020}, rely on overly restrictive assumptions. For example, BERTScore assumes a one-to-one token mapping between reference and predicted sequences. We demonstrate that relaxing this assumption leads to improvements, even without employing advanced deep contextualized embeddings like BERT \cite{Devlin2019}.

To address the aforementioned challenges, we introduce Word Rotator's Distance Score (WRDScore), a novel NLG metric. WRDScore features a normalized output value between 0 and 1, providing more meaningful and interpretable results. This normalized score offers a clearer indication of an NLG model's performance relative to the optimal case. Additionally, WRDScore is versatile and can utilize various word embeddings, including both contextualized (ELMo, BERT) and non-contextualized (Word2Vec, GloVe) options. By employing the optimal transport formulation \cite{Peyre2019-ot}, WRDScore enables soft assignment between compared sentences, incorporating context into measurements while reducing dependence on the context encoded in individual word embeddings. This inherent property allows WRDScore to surpass other advanced NLG metrics, such as BERTScore, even without relying on contextualized word embeddings. In this regard, WRDScore can be viewed as a generalization and relaxation of BERTScore.

\section{Related work}

\subsection{ROUGE}
The ROUGE metric \cite{Lin2004}, or Recall-Oriented Understudy for Gisting Evaluation, is the most straightforward and widely adopted approach for comparing two text sentences. It assesses the direct overlap between the token sets of the input sentences. The most popular variant of ROUGE is ROUGE-N, typically with N = 1 or 2. Specifically, ROUGE-N calculates the number of matching n-grams between the model-generated text and the human-produced reference sentence. Formally, ROUGE-N is defined as a recall-oriented measure:
$$
\text{ROUGE-N} = \frac{ \sum_{S \in R } \sum_{ gram_n \in S } Count_{match} (gram_n) }{ \sum_{ S \in R } \sum_{ gram_n \in S } Count(gram_n) },
$$
where $R$ refers to the set of reference summaries, $N$ stands for the length of the n-gram ($gram_n$), and $Count_{match}(gram_n)$ is the maximum number of n-grams co-occurring in the candidate summary and the set of reference summaries. A closely related measure, BLEU \cite{Papineni2002-bleu}, used in automatic evaluation of machine translation, is a precision-based measure. BLEU measures how well a candidate translation matches a set of reference translations by counting the percentage of n-grams in the candidate translation overlapping with the references.

\subsection{BERTScore}
BERTScore \cite{Zhang2020} uses BERT \cite{Devlin2019} contextualized embeddings to compute the pairwise cosine similarity of each reference token $i$ with each candidate token $j$. Then, the precision and recall are calculated through the maximum-based hard matching of the tokens in candidate (predicted) sentence $p$ to the tokens of reference sentence $r$ as follows:
$$
BERTScore = F_{BERT} = 2 \frac{PR_{BERT} \cdot RC_{BERT}}{PR_{BERT} + RC_{BERT}}
$$
$$
PR_{BERT} = \frac{1}{|p|} \sum_{j \in p} \max_{i \in r} \vec{i}^T \vec{j}
$$
$$
RC_{BERT} = \frac{1}{|r|} \sum_{i \in r} \max_{j \in p} \vec{i}^T \vec{j}
$$

BERTScore uses a hard alignment of tokens to compute recall, as illustrated in Figure~\ref{fig:hard_alignment}:
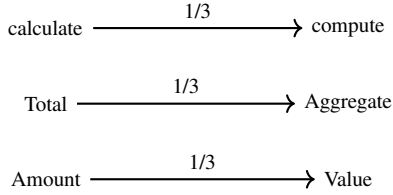
\begin{figure}[h]
	\centering
	\begin{tikzpicture}[every node/.style={font=\footnotesize}]
		\node (calculate) at (0,0) {calculate};
		\node (total) at (0,-1) {Total};
		\node (amount) at (0,-2) {Amount};
		
		\node (compute) at (4,0) {compute};
		\node (aggregate) at (4,-1) {Aggregate};
		\node (value) at (4,-2) {Value};
		
		\draw[->, thick] (calculate) -- (compute) node[midway, above] {1/3};
		\draw[->, thick] (total) -- (aggregate) node[midway, above] {1/3};
		\draw[->, thick] (amount) -- (value) node[midway, above] {1/3};
	\end{tikzpicture}
	\caption{An example of a hard alignment between two word sequences (Java method names) ``calculateTotalAmount'' (reference) and ``computeAggregateValue'' (predicted).}
	\label{fig:hard_alignment}
\end{figure}

Instead, a soft token alignment is more desirable for recall calculation (Figure~\ref{fig:soft_alignment}):
\begin{figure}[h]
	\centering
	\begin{tikzpicture}[every node/.style={font=\footnotesize}]
		\node (calculate) at (0,0) {calculate};
		\node (total) at (0,-2) {Total};
		\node (amount) at (0,-4) {Amount};
		
		\node (compute) at (5,0) {compute};
		\node (aggregate) at (5,-2) {Aggregate};
		\node (value) at (5,-4) {Value};
		
		\draw[->, thick] (calculate) -- (compute) node[near start, above, sloped] {7/30};
		\draw[->, thick] (calculate) -- (aggregate) node[near start, above, sloped] {1/15};
		\draw[->, thick] (calculate) -- (value) node[near start, above, sloped] {1/30};
		
		\draw[->, thick] (total) -- (compute) node[near start, above, sloped] {1/15};
		\draw[->, thick] (total) -- (aggregate) node[near start, above, sloped] {7/30};
		\draw[->, thick] (total) -- (value) node[near start, above, sloped] {1/30};
		
		\draw[->, thick] (amount) -- (compute) node[near start, above, sloped] {1/30};
		\draw[->, thick] (amount) -- (aggregate) node[near start, above, sloped] {1/15};
		\draw[->, thick] (amount) -- (value) node[near start, above, sloped] {7/30};
	\end{tikzpicture}
	\caption{An example of a soft alignment between two word sequences (Java method names) ``calculateTotalAmount'' (reference) and ``computeAggregateValue'' (predicted).}
	\label{fig:soft_alignment}
\end{figure}
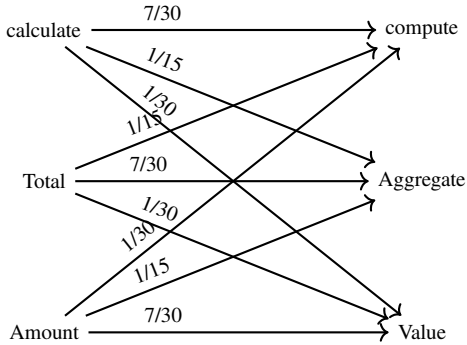

\subsection{MoverScore}
MoverScore \cite{Zhao2019} was inspired by using the Word Mover's Distance \cite{Kusner2015} (WMD) and takes the value equal to the cost of the optimal transportation plan between the reference and predicted sentences when they are viewed as probability distributions. MoverScore is able to use Word2Vec, BERT, or ELMo to define word embeddings. Also, MoverScore assigns weights for the defined word vectors via TF-IDF. Word embeddings can be n-grams.

It was observed that MoverScore \cite{Zhao2019} only marginally outperforms BERTScore when measured on a variety of tasks, such as machine translation, image captioning, and dialog response generation. Experiments also show that MoverScore outperforms both asymmetric factors: $PR_{BERT}$ and $RC_{BERT}$ \cite{Zhao2019}. However, when combined via harmonic mean, BERTScore matches MoverScore's performance. We hypothesize that BERT's contextualized embeddings, which encode sentence-level information into each word vector, soften BERTScore's hard alignments, leading to this phenomenon.

BERTScore is a precision-recall-based metric. Thus, it is naturally normalized and provides a better interpretation of an NLG model's performance. This interpretation is useful when the model's predictions either lack precision (i.e., do not guess true tokens precisely) or recall (do not cover true sentences well enough), or the evaluation dataset is highly imbalanced. The vast majority of embedding-based NLG metrics proposed in the literature neither follow the precision-recall framework, nor output normalized scores. In contrast, we propose a natural soft-alignment-based approach to measure precision and recall from the optimal transport cost value between the reference and predicted sentences. To the best of our knowledge, no other existing embedding-based metric provides such a property. Also, contrary to MoverScore, the word vectors used in our algorithm are pre-normalized and weighted by either the original vector norms, TF-IDF scores, or a combination of both. This enables to consider the pairwise cosine similarity between word vectors instead of Euclidean distance used in MoverScore.

\section{Proposed Algorithm}

Suppose the reference and predicted token sequences are given by $r = \{ r_1, \ldots, r_n \}$ and $p = \{ p_1, \ldots, p_m \}$, respectively. Then, the usual precision and recall can be calculated as follows:
$$
\text{True Positive} \ TP = | r \cap p |
$$
$$
\text{False Positive} \ FP = | p \setminus r |
$$
$$
\text{False Negative} \ FN = | r \setminus p |
$$
$$
\text{Precision} \ PR = \frac{TP}{TP + FP}
$$
$$
\text{Recall} \ RC = \frac{TP}{TP + FN}
$$
$$
\text{F1 Score} \ F1 = \frac{2 \cdot PR \cdot RC}{PR + RC}
$$

If we associate each token with its one-hot embedding vector, i.e.  $\vec{t_i} := (0, \ldots, 0, 1, 0, \ldots, 0)$, where $1$ is placed at the $i^{th}$ position, then the above precision and recall formulas are just special cases of the more general ones:
$$
	PR = \frac{1}{| p |} \sum_{j \in p} \max_{i \in r} \vec{r_i}^T \cdot \vec{p_j}
$$
$$
	RC = \frac{1}{| r |} \sum_{i \in r} \max_{j \in p} \vec{r_i}^T \cdot \vec{p_j}
$$

Clearly, more advanced embedding algorithms can be utilized (e.g., from the co-occurrence matrix, Word2Vec, ELMo, BERT, and so on). Therefore, suppose the reference and predicted sequences are now represented via normalized embeddings of an advanced embedding mechanism:
$$
r = \{ \vec{r_1}, \ldots, \vec{r_n} \}
$$
$$
p = \{ \vec{p_1}, \ldots, \vec{p_m} \}
$$

Also, we assign probability weights:
$$
W^r = \{ w^r_1, \ldots, w^r_n \} \in \Sigma_n
$$
$$
W^p = \{ w^p_1, \ldots, w^p_m \} \in \Sigma_m
$$
The norms of the unnormalized token embeddings or TF-IDF scores are used as weights, reflecting token importance.

Consider the Kantorovich optimal transport formulation \cite{Peyre2019-ot}:
\begin{align}
    \min_{P \in \mathbb{R}^{n \times m}} & \sum_{i, j} P_{ij} \cdot d_{ij} \label{eq:ot_problem_form} \\
    P \cdot \mathbf{1} = W^r&, \quad \mathbf{1}^T \cdot P = W^p \label{eq:ot_problem_contr}
\end{align}

Then, the Word Rotator's Distance (WRD) \cite{Yokoi2020} is:
$$
WRD((r, W^r), (p, W^p)) = \sum_{i, j} P^*_{ij} \cdot d_{ij},
$$
where $\{P^*_{ij}\}$ represents the optimal flow matrix obtained as a solution of the Kantorovich problem \eqref{eq:ot_problem_form}-\eqref{eq:ot_problem_contr}.

It is natural to define the precision as the hard token-wise inclusion of the predicted sequence into the reference one:
\begin{equation}
PR_{WRDScore} = \sum_{j \in p} W^p_j \cdot \max_{i \in r} \vec{r_i}^T \cdot \vec{p_j} \label{eq:wrd_score_pr}
\end{equation}

Also, we naturally define the recall as the measure of how easily the reference sequence distribution can be softly and optimally transported into the predicted one, or covered by it:
$$
RC_{WRDScore} = 1 - \sum_{i, j} P^*_{ij} \cdot d_{ij}
$$
The use of cosine distance in the Kantorovich optimization problem \eqref{eq:ot_problem_form}-\eqref{eq:ot_problem_contr} and cosine similarity in \eqref{eq:wrd_score_pr} ensures that the similarity scores between pairs of tokens naturally range between 0 and 1. In contrast, using Euclidean distance, as in MoverScore, would require an artificial bounding procedure to define the new precision and recall formulas, and the optimality of this procedure would need to be investigated.

Finally, Word Rotator's Distance Score (WRDScore) is defined as:
$$
WRDScore(r, p) = \frac{2 \cdot PR_{WRDScore} \cdot RC_{WRDScore}}{PR_{WRDScore} + RC_{WRDScore}}
$$

\section{Experiments}

To demonstrate the efficiency of the proposed metric, we conducted experiments using the Java large dataset of Java code \footnote{https://github.com/tech-srl/code2seq}, performing a human evaluation of UniXcoder's \cite{Guo2022} output for method name prediction.

We considered three sets of method name pairs consisting of 50 samples each, where the samples were obtained from the results generated by the UniXcoder model fine-tuned on the above-mentioned dataset of Java methods. The samples were selected as follows:

\begin{itemize}
\item 10 random samples from the entire dataset;
\item 30 samples drawn uniformly at random from method name pairs, divided into three WRDScore ranges: $[0, 0.2]$, $[0.2, 0.5]$, and $[0.5, 0.8]$;
\item 10 samples from pairs with WRDScore in $[0.8, 1.0]$ and ROUGE-1 $< 0.5$ (to avoid sampling identical pairs).
\end{itemize}

Token embeddings were trained using Word2Vec on the method names with the context window size 2. The obtained three sets of 50 method name pairs were evaluated by 9 experienced Java programmers, who assigned semantic similarity scores (from 0 to 1). The average scores of 3 different responders for each set were used as ground truth.

We computed the deviation of ROUGE-1 and WRDScore metric values from the ground truth using MSE and MAE. The results are shown in Table~\ref{tab:exp_results}.

\begin{table}[ht]
	\begin{center}
	\caption{Mean deviation from ground truth (human scores) for the compared metrics}
	\begin{tabular}{|c|ccc|}
		\hline
		& ROUGE-1 & BERTScore & WRDScore \\
		\hline
		MSE & 0.0817 & 0.2769 & \textbf{0.0678} \\
		MAE & 0.2213 & 0.4312 & \textbf{0.1891} \\
		\hline
	\end{tabular}
	\label{tab:exp_results}
	\end{center}
\end{table}

The results were validated by T-tests for the means of two independent samples with a high statistical significance (p-value $\ll$ 0.05). WRDScore showed 17.01\% improvement in MSE and 14.55\% in MAE over ROUGE-1.

\section{Reproducibility Package}

The reproducibility package for this article, including the curated human evaluation dataset, is available at GitHub \footnote{https://github.com/rmusab/wrd-score}.

\section{Conclusion}

In this work, we proposed WRDScore, which possesses the following unique properties:
\begin{itemize}
	\item A normalized, precision-recall-based metric, providing better interpretation of NLG models' performance, as well as being useful for imbalanced datasets and models lacking either precision or recall;
	
	\item Naturally leveraging a soft alignment of pre-normalized word vectors weighted by their original norms or TF-IDF coefficients through the optimal transport problem (unlike BERTScore);
	
	\item Using cosine similarity instead of Euclidean distance (unlike MoverScore).
\end{itemize}

We experimentally tested the new metric using a human evaluation dataset. The results clearly demonstrate the superiority of WRDScore over existing metrics.

\section{Acknowledgement}

This research was funded by the Science Committee of the Ministry of Science and Higher Education of the Republic of Kazakhstan (grant no. BR21882268).

\bibliographystyle{IEEEtran}
\bibliography{references}

\end{document}